# Short Communication on QUIST: A Quick Clustering Algorithm

Sherenaz W. Al-Haj Baddar
Department of Computer Science, KASIT
The University of Jordan
Amman, 11942 Jordan
s.baddar@ju.edu.jo

**Abstract**
In this short communication we introduce the **quick** clu**st**ering algorithm (QUIST), an efficient hierarchical clustering algorithm based on sorting. QUIST is a poly-logarithmic divisive clustering algorithm that does not assume the number of clusters, and/or the cluster size to be known ahead of time. It is also insensitive to the original ordering of the input.

## Introduction

Clustering is an unsupervised learning mechanism which aims at grouping similar instances in groups, where no examples exist to help identify the correct cluster to which a given node belongs [1]. Hierarchical clustering algorithms build a structure as they assign instances to their corresponding clusters using a proximity measure, and can be either agglomerative or divisive [2]. In agglomerative clustering, each instance is assumed to be a cluster on its own, then, clusters get grouped together. Divisive clustering, on the other hand, considers the initial set of data instances one big cluster, and starts splitting it to form sub-clusters, until a pre-defined termination condition is met. Our clustering algorithm, QUIST, is a divisive univariate clustering algorithm. In this context, let us denote the initial cluster, in QUIST, by $C$, and also denote the value associated with data instance $i$ by $v_i$.

## The QUIST Algorithm

QUIST begins by sorting all input data instances in its initial cluster $C$, according to their univariate values in non-decreasing order. Then, it starts splitting $C$ into smaller sub-clusters, until either one of the following conditions holds, whichever happens first:

1. The instances within each sub-cluster, $c$, are too similar to be divided any further, or
2. The number of clusters hits an optional upper bound provided by the user. If such bound is not provided by the user, it is assume to be equal to the number of input instances, $|C|$

Additionally, splitting a given cluster stops when it reaches a minimum cluster size per the user's choice.

To decide whether or not instances within a cluster, $c$, are similar enough to stop splitting it, QUIST calculates the "spreadness" metric denoted by $\Psi$, such that the spreadness of $c$, denoted by $\Psi_c$ is

$$\Psi_C = \begin{cases} \frac{\sigma_C}{(Q3_c - Q1_c)}, & Q3_c \neq Q1_c \\ 0, & Q3_c = Q1_c \end{cases} \tag{1}$$

where, $\sigma_c$ denotes the standard deviation of the values $v_i$ in a given cluster $c$, while $Q1_c$ and $Q3_c$ denote the first and third quartiles of these values, respectively. The spreadness metric aims at expressing the scatterness of the values within a given cluster. If values $v_i$ are laid out in the form of a boxplot, then $(Q3_c - Q1_c)$ denotes the length of the boxplot. As a large standard deviation with respect to the length of the boxplot implies more "scatterness" of the data, QUIST considers that a cluster with $\Psi_C$ greater than one needs further splitting. If $Q3_c$ is equal to $Q1_c$, then $\Psi_c$ is considered to be zero, as this implies that the instances in the cluster are potentially close enough to form one cluster . Figure 1 depicts how QUIST algorithm operates.

**QUIST algorithm**

Input:

- $C$, denotes the set of all univariate instances to be clustered
- $K$, denotes the maximum number of clusters the user wishes to have [set to $|C|$, if not provided by the user].
- $s$, denotes the minimum cluster size the user wishes to have [set to one, if not provided by the user].

Output:

- $L$, denotes the set of clusters generated by QUIST

Begin

1. $L \leftarrow \Phi$
2. Sort the values associated with the instances in $C$ to identify their first, second, and third quartiles denoted by $Q1_C$ , $Q2_C$ , and $Q3_C$ respectively.
3. Calculate the spreadness of cluster $C$, denoted by $\Psi_C$
4. If $\Psi_C \leq 1$ or $|C| \leq s$ or $K = 1$, then
   4.1 $L \leftarrow C$
   4.2 Label $L$ as "done"
   4.3 Stop
5. $c \leftarrow C$
6. If $|L| < K$, then
   6.1 Split $c$ into two clusters, denoted by $c_1$ and $c_2$ , such that
   $$c_1 = \{i\ ,\ \forall\, i\ \in c_1\ \ v_i \leq\ Q2_c\ \}$$
   and
   $$c_2 = \{i\ ,\ \forall\, i\ \in c_2\ \ v_i >\ Q2_c\ \}$$
   6.2 If $L \neq \Phi$ then, remove $c$ from $L$
   6.3 If $c_1 = \Phi$, then
      6.3.1 Label $c_2$ as "done"
      6.3.2 Add $c_2$ to $L$
   6.4 Else, if $c_2 = \Phi$, then
      6.4.1 Label $c_1$ as "done"
      6.4.2 Add $c_1$ to $L$
   6.5 Else, Add $c_1$ and $c_2$ to $L$, initially unlabeled
   6.6 For each unlabeled cluster, $c$, in $L$ do
      6.6.1 If $\Psi_c \leq 1$ or $|c| \leq s$, then label $c$ as "done"
      6.6.2 Else, label $c$ as "undone"
   6.7 If there are no "undone" clusters in $L$, then stop
   6.8 Among all clusters labeled "undone" in $L$, select cluster, $c$, with max $\Psi_C$.
   6.9 Go to step 6 and repeat.
7. Else,
   7.1 Label all "undone" clusters in $L$ as "done"
   7.2 Stop.

End

**Fig. 1. The QUIST Clustering Algorithm**

**Remarks on QUIST clustering algorithm**

*The Efficiency of QUIST*

If we denote the input size $|C|$ by $N$, then we can show that QUIST requires $O(N\log N)$ operations to achieve its task. First of all, QUIST sorts all values associated with the initial cluster $C$, consuming $O(N\log N)$ using an optimal sorting algorithm. Then, QUIST splits the cluster with maximum spreadness to examine it further, during each clustering iteration. As there are at most $N$ sorted values in a given cluster, and no more than $K$ clustering iterations, QUIST needs $O(KN)$ operations to finish clustering. Thus, and given that $K$ is, typically, a constant, QUIST runs in $O(N\log N)$ operations. Therefore, QUIST is more efficient than many hierarchical clustering algorithms with quadratic complexities or even worse [3, 4].

*The Sensitivity of QUIST*

Many clustering algorithms are known to be sensitive to input ordering, which implies that they produce different clustering for the same input set, should it be presented differently. Examples include the well-known k-means, BIRCH, and Self-Organizing Maps (SOM) algorithms to mention few [5-7]. QUIST, on the other hand, is insensitive to the initial ordering of the data; thus, it produces the same clustering for a given input, regardless of input ordering.

*Indivisible Clusters in QUIST*

It is possible that a given cluster $C$ is indivisible in QUIST, that is, when QUIST tries to split the cluster with respect to its second quartile $Q2_C$, it might end up with either $c_1$ or $c_2$ being empty, as illustrated in steps 6.3 and 6.4 in the algorithm in Fig. 1. In either case, the resulting non-empty cluster is labeled "done", as it cannot be split any further according to the spreadness metric. This implies that a cluster whose instances are not close enough could be generated by QUIST. However, this condition will shift the algorithm's effort to the next cluster which could possibly benefit from further clustering, and guarantees the termination of QUIST.

**Acknowledgements**

The author wishes to thank Dr. Eva Riccomagno (University of Genoa) for her instrumental insights which helped formulate the spreadness metric. She also wishes to thank Dr. Mauro Migliardi (University of Padua), Dr. Alessio Merlo (University of Genoa), and Dr. Davide Anguita (University of Genoa), for motivating the development of QUIST.